\documentclass[paperpaper, 10 pt, conference]{ieeeconf}
\IEEEoverridecommandlockouts
\usepackage{tikz}
\usepackage{textcomp}
\usepackage{hyperref}
\usepackage{flushend}

\usepackage{url}
\usepackage{caption}
\usepackage{xcolor}
\usepackage{lipsum}
\usepackage{cite}
\usepackage{amsmath,amssymb,amsfonts}
\usepackage{hyperref}
\usepackage{algorithmic}
\usepackage{graphicx}
\usepackage{siunitx}
\usepackage[utf8]{inputenc}
\usepackage{textcomp}
\usepackage{amsmath}
\usepackage{amsfonts}
\usepackage{commath}
\usepackage{float}
\let\labelindent\relax
\usepackage{enumitem}
\usepackage{epsfig}
\usepackage{epstopdf}
\usepackage[fleqn]{nccmath}
\usepackage[linesnumbered, ruled, vlined]{algorithm2e}
\usepackage{gensymb}

\def\BibTeX{{\rm B\kern-.05em{\sc i\kern-.025em b}\kern-.08em
    T\kern-.1667em\lower.7ex\hbox{E}\kern-.125emX}}

\newcommand*{\rom}[1]{\expandafter\@slowromancap\romannumeral #1@}
\makeatother

\SetCommentSty{mycommfont}
\newcolumntype{P}[1]{>{\centering\arraybackslash}p{#1}}
\newcolumntype{M}[1]{>{\centering\arraybackslash}m{#1}}

\IEEEoverridecommandlockouts

\begin{document}

\raggedbottom
\title{\LARGE \bf OptiState: State Estimation of Legged Robots using Gated Networks with Transformer-based Vision and Kalman Filtering}

\author{Alexander Schperberg$^{1}$, Yusuke Tanaka$^{1}$, Saviz Mowlavi$^{2}$, Feng Xu$^{1}$, \\ Bharathan Balaji$^{3}$, Dennis Hong$^{1}$
\thanks{$^{1}$A. Schperberg, Y. Tanaka, F. Xu, and D. Hong are with the Robotics and Mechanisms Laboratory, Department of Mechanical and Aerospace Engineering, UCLA, Los Angeles, CA, 90024, USA. {\tt\small \{aschperberg28, yusuketanaka, xufengmax,dennishong\}@ucla.edu}}
\thanks{$^{2}$S. Mowlavi is with the Mitsubishi Electric Research Laboratories (MERL), Cambridge, MA, 02139, USA. {\tt\small \{mowlavi\}@merl.com}}
\thanks{$^{3}$B. Balaji is with Amazon Science, Seattle, WA, 98109, USA. {\tt\small \{bhabalaj\}@amazon.com}.
Work not related to Amazon Science.}}

\maketitle

\global\csname @topnum\endcsname 0
\global\csname @botnum\endcsname 0
\begin{abstract}
State estimation for legged robots is challenging due to their highly dynamic motion and limitations imposed by sensor accuracy. By integrating Kalman filtering, optimization, and learning-based modalities, we propose a hybrid solution that combines proprioception and exteroceptive information for estimating the state of the robot's trunk. Leveraging joint encoder and IMU measurements, our Kalman filter is enhanced through a single-rigid body model that incorporates ground reaction force control outputs from convex Model Predictive Control optimization. The estimation is further refined through Gated Recurrent Units, which also considers semantic insights and robot height from a Vision Transformer autoencoder applied on depth images. This framework not only furnishes accurate robot state estimates, including uncertainty evaluations, but can minimize the nonlinear errors that arise from sensor measurements and model simplifications through learning. The proposed methodology is evaluated in hardware using a quadruped robot on various terrains, yielding a 65\% improvement on the Root Mean Squared Error compared to our VIO SLAM baseline. Code example: \texttt{https://github.com/AlexS28/OptiState}

\end{abstract}


\section{Introduction}
Autonomous legged locomotion typically employs sophisticated planners and/or controllers, whose success and stability are directly dependent on an accurate state estimation system. 
However, state estimation for legged robots is difficult because dynamic motion produced by footsteps can cause camera motion blur, making vision data less reliable. Due to constantly making and breaking contact, along with slipping, kinematic information can be error prone as well, particularly in environments with large disturbances (i.e., obstacles) or compliant surfaces.
Because legged robots produce dynamic motion which demands high frequency control, low computational cost then becomes a necessary design requirement for state estimation systems.
\begin{figure}[!t]
    \centering
    \includegraphics[width=0.95\columnwidth]{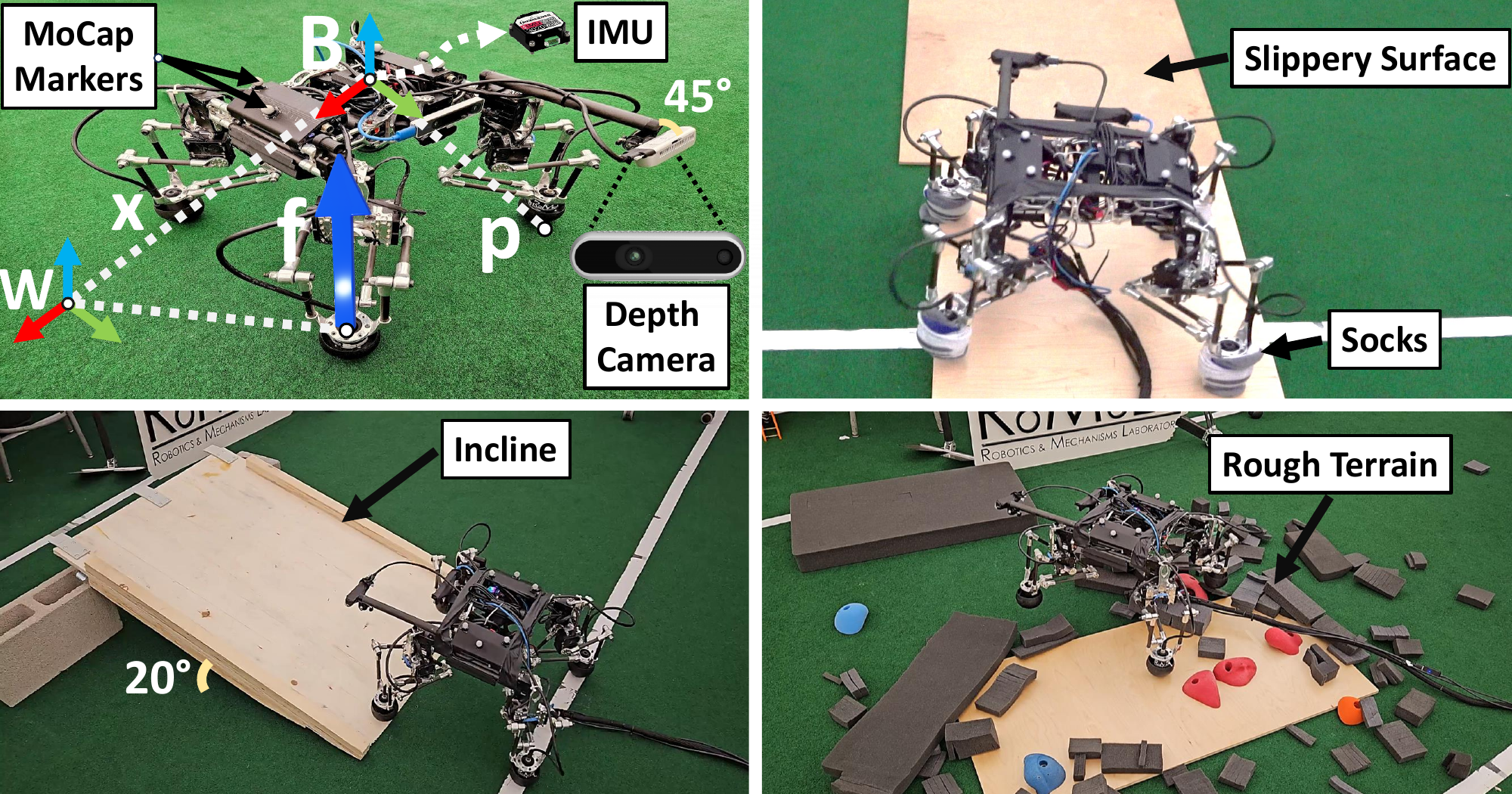}
    \captionsetup{font=small} 
    \caption{Top left shows the attached body frame (B), and world frame (W). The trunk state $\mathbf{x}$ is in the world frame, while footstep positions $\mathbf{p}$ is relative to the body. Ground reaction forces (blue arrow) from our MPC control policy are given by $\mathbf{f}$, also in world frame. We verify our algorithm, OptiState, on slippery surfaces (top right), incline (bottom left), and rough terrain (bottom right).} 
    \label{intro}
\end{figure}
Overall, state estimation methods on legged robots can be subdivided into several design approaches. For example, methods that employ Kalman filtering on either proprioception alone \cite{fallon2019}, combining proprioception with exteroceptive information \cite{DurrantWhyte2006}, using optimization \cite{Xinjilefu2014} or neural networks \cite{Brossard2020} on some part of the overall estimation framework, and/or with the addition of factor graphs \cite{buchanan2021learning}. 

In this work, we take a hybrid approach that combines model-based Kalman filtering, optimization, and learning methods for state estimation using proprioception and exteroceptive information. The motivation of this approach is to facilitate fast nonlinear modeling in Visual-Inertial Odometry (VIO) estimation while addressing the limitations of pure model-based methods in capturing error from inaccurate state or measurement models, and achieve generalization in learning by incorporating domain knowledge.

Specifically, we use a Kalman filter that takes as input joint encoder and IMU measurements,
and employ a single-rigid body model to propagate states by reusing the ground reaction force control outputs from a convex Model Predictive Control (MPC) optimization \cite{forceMPC}. The output of this Kalman filter, along with the state input history over a receding time horizon, is then fed as input to a Gated Recurrent Unit neural network (GRU).

Additionally, the GRU uses as input the latent space representation of depth images through a Vision Transformer (ViT) \cite{MASKED_ViT}, which helps provide information not only on robot height, but can infer semantic information about its environment to help the estimator. After an offline training phase using a loss function which compares its prediction with the ground truth state output provided by a motion capture system, the goal of the GRU is to update (or correct) the Kalman filter state output to help alleviate the nonlinearities that may exist from noisy measurements and inaccuracies of our single-rigid body model. Using this hybrid approach, we will demonstrate the improved generalization capabilities of our GRU. Specifically, our GRU leverages the Kalman filter's output knowledge and image latent space vector, while remaining robust in predicting state components even in scenarios where our model-based Kalman filter assumptions break down, such as assuming non-slipping conditions and small angle approximations in roll and pitch, see Sec. \ref{model}.

This paper presents the following \textbf{contributions}:
\begin{enumerate}
\item We combine a model-based Kalman filter with the ability of learning nonlinearities through a GRU and ViT. Our estimator provides both the robot's trunk state and the associated predictive uncertainty. 
\item For our Kalman filter, we consider joint encoder and IMU measurements, and reuse the control outputs from a convex MPC for the filter's system model.
\item We demonstrate our state estimator on hardware using a quadrupedal robot on various terrains and demonstrate a 65\% improvement against a state-of-the-art VIO Simultaneous Localization and Mapping (SLAM) baseline \cite{hausamann2021evaluation} using the Root Mean Squared Error (RMSE) as our validation metric.
\end{enumerate}

\subsection{Related Works}
In \cite{fallon2019}, an Extended Kalman Filter (EKF) approach was used on the Atlas bipedal robot, where the measurements include the velocity derived from the integration of leg odometry. An EKF was also employed in \cite{agarwal2023} to fuse leg kinematics and IMU information, and an a Unscented Kalman Filter (UKF) in \cite{Bloesch2013} to better handle the nonlinearities of slippery terrain. However, these methods suffer from linearization errors \cite{fink2020}. Additionally, these works do not consider vision or other exteroceptive information which may impose drift over long term operation, which may be avoided in areas cameras can be applied. 

One solution is to employ both proprioception and exteroceptive information through SLAM \cite{DurrantWhyte2006, Bailey2006}, visual odometry \cite{Scaramuzza2011, Fraundorfer2012}, or more currently, VIO SLAM \cite{Concha2016, pronto, cerberus}. However, while these works benefit from the simplicity and computational efficiency of Kalman filter methods, they are still limited by their requirement to approximate highly nonlinear systems and rely on sufficient initial conditions. In our case, employing a simple Kalman filter to estimate the robot's state, when utilized as input for the GRU, enhances the GRU's performance, especially concerning velocity components in unseen datasets (see Sec. \ref{Results}).

Due to these issues associated with methods that rely purely on Kalman filtering, some works, such as in \cite{Xinjilefu2014}, employ state estimation through Quadradic Programming (QP), which considers both the modeling and measurement error within its cost function. However, while their QP works well in simulation, where their measurement noise appears relatively low, they do not compare their algorithm against ground truth on the real robot (e.g., using a motion capture system). Instead, they only compare their QP against a Kalman filter on the real robot without ground truth. 
Although our use of QP differs from the work in \cite{Xinjilefu2014}, using QP for state estimation is a realistic option due to the recent advancement of computers. In our case, we use a convex MPC QP program \cite{forceMPC} which computes ground reaction forces that actuate a robot to help follow a user's reference trajectory. Thus, from their model formulation, these forces embed information of the user's reference trajectory (e.g., velocity command), and also the previous state estimator output. These forces are then reused as part of our Kalman filter's system model to propagate the state to the next time step.   

In this work, we further correct any errors due to nonlinearities that occur from the model and measurements through a GRU. Using learning to correct for model errors is validated by several works. For example, using IMU-only state estimation on pedestrian motion \cite{Brossard2020} using a neural network, learning model/measurement errors through a Recurrent Neural Network (RNN) by removing bias errors and learning IMU noise parameters \cite{Zhang2021}, and combing a UKF with a neural network to learn the residual errors between a dynamic model and the real robot \cite{schperberg2023realtosim}. However, many of these works, such as \cite{Zhang2021,schperberg2023realtosim,satorras2019combining,saber}, focus more on smoothing, imputation, or error prediction, rather than providing an end-to-end state estimation system.

To incorporate a history of states and measurements for estimating the current state can be done through factor graphs. For example, \cite{buchanan2021learning} uses factor graphs to combine motion priors with kinematics based on historical state and measurement data. While contact state estimation can be useful, it's not always required as shown in \cite{Wisth2019}, which uses factor graphs with inertial, leg, and visual odometry on soft and slippery surfaces without requiring force/torque sensors. However, factor graphs come with certain limitations. They necessitate an accurate dynamic model and encounter scalability challenges when dealing with significantly expanded input state spaces, which can result in substantial computational overhead. We address these issues using a GRU, which allows us to employ a greater number of state variables while also considering time-series inputs without significant computational effort. Additionally, instead of using the entire depth image as input, we use a ViT to down-scale the image into a smaller latent space, making the overall GRU faster to train. 

Perhaps the work most similar to ours can be found in \cite{Revach2022}, which learns the Kalman gain parameters through a GRU for state estimation. However, there are several distinct differences besides the methodology itself. For one, we explicitly learn not only the state estimate but also the potential error associated with the networks' prediction at each time step. This serves as an indicator for situations involving high prediction errors where we can still utilize the Kalman filter's output, as it operates independently of any learning components. Lastly, we incorporate vision as part of the error prediction, and apply our methods on complex hardware online. To our knowledge, we are the first to develop an end-to-end nonlinear state estimation system for legged robots, which directly employs the output of a machine learning (ML) model while enhanced by the integration of the Kalman filter and depth encoder into its input space. We outperform a state-of-the-art VIO SLAM solution with our approach.  

\section{Methods}
\label{methods}
The rest of the paper is organized as follows: an introduction to our overall framework, called OptiState, is given in Sec. \ref{problem_definition}, the measurements and model we employ in Sec. \ref{measurement} and Sec. \ref{model} respectively, the Kalman filter equations in Sec. \ref{kalman_filter}, and the GRU to update our Kalman filter estimation and uncertainty in its prediction as well as the ViT to encode the depth image into latent space in Sec. \ref{learning}.

\begin{figure}[!t]
    \centering    \includegraphics[width=0.99\columnwidth]{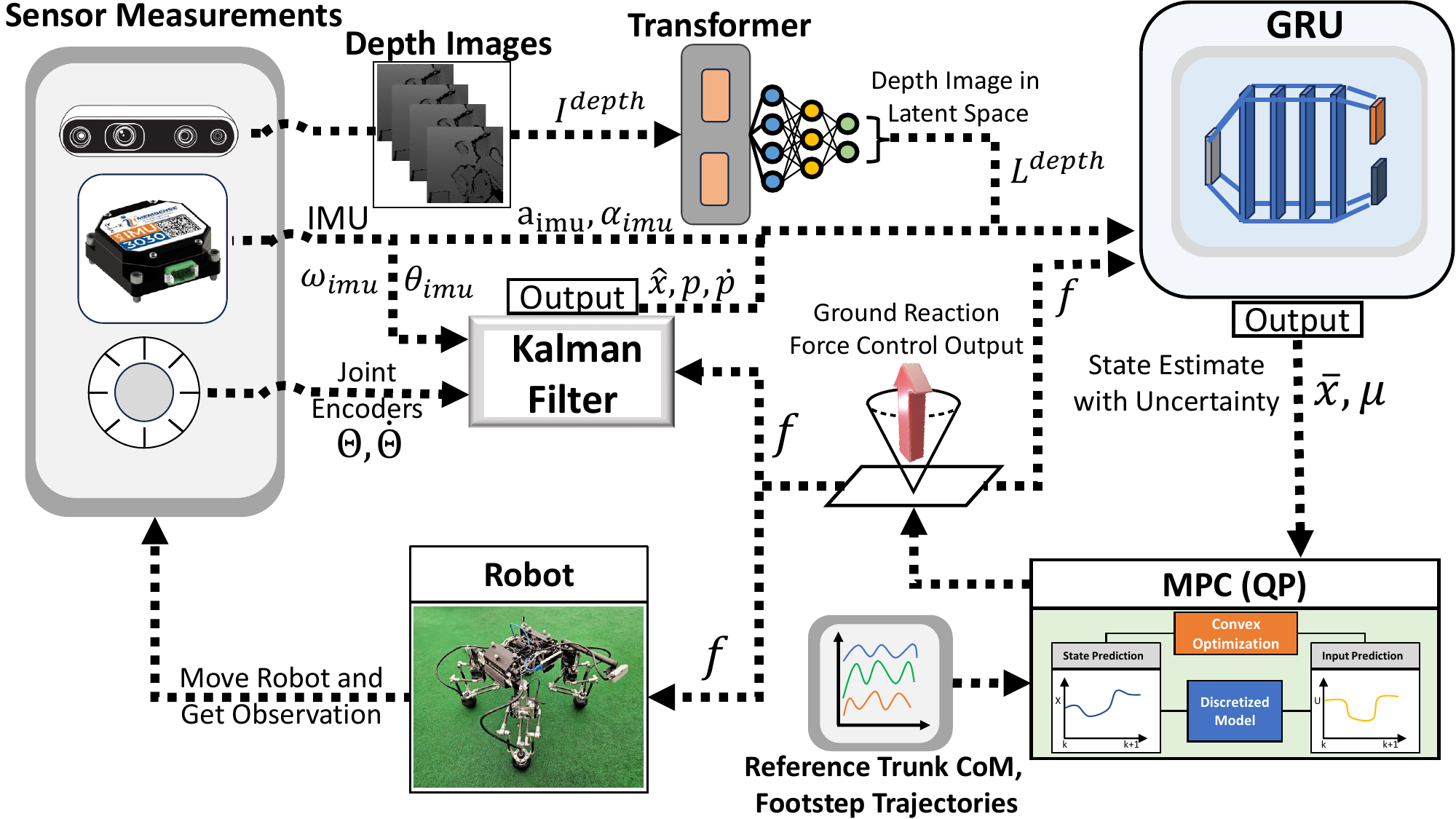}
    \captionsetup{font=small} 
    \caption{Overall state estimation architecture as described in Sec. \ref{methods}. } 
    \label{method_architecture}
\end{figure}

\subsection{Problem Definition}
\label{problem_definition}
Our state estimation framework estimates the trunk state (robot's CoM) orientation, position, angular, and linear velocity in the world frame. We assume access to a 6-axis IMU, depth camera, joint encoder position and velocity, and during the training procedure of the GRU, access to the ground truth (i.e., motion capture system). The variables used and their size are compiled in Table \ref{tab:my_label}. Our framework can be summarized in two parts: first, a Kalman filter is formulated that provides an initial estimate of the robot's trunk state:
\begin{align}
    \label{eq:kalman}
    \hat{\mathbf{x}}_{k} = \text{KF}(\boldsymbol{\Theta}_{k}, \boldsymbol{\dot{\Theta}}_{k}, \boldsymbol{\theta}^{\rm imu}_{k}, \boldsymbol{\omega}^{\rm imu}_{k}, n^{c}_{k}, \mathbf{f}_{k},\hat{\mathbf{x}}_{k-1})
\end{align}
where $\boldsymbol{\Theta}_{k}$, $\dot{\boldsymbol{\Theta}}_{k}$ are the joint encoder positions and velocities for all feet, $\boldsymbol{\theta}_{k}^{\rm imu}$, $\boldsymbol{\omega}_{k}^{\rm imu}$ are the Euler orientation (converted from raw quaternion values) and angular velocity from the IMU, $n_{k}^{c}$ is the number of feet currently in contact with the ground given by reference contact matrix $\mathbf{C}$, $\mathbf{f}_{k}$ are the ground reaction forces generated by an MPC controller formulated identically to \cite{forceMPC} (model described in Sec. \ref{model}), $k$ is the current time step, and the output of the filter is $\hat{\mathbf{x}}_{k}$ or the trunk state of the robot which includes the following order of components: heading angle in roll, pitch, and yaw, position in $x$, $y$, and $z$, angular velocity in roll, pitch, and yaw, and linear velocity in $x$, $y$, and $z$, or $\hat{\mathbf{x}}_{k}=[\theta_{x},\theta_{y},\theta_{z},r_x,r_y,r_z,\omega_x,\omega_y,\omega_z,v_x,v_y,v_z]^\top$.

Note, because many legged robots \cite{forceMPC} use ground reaction force control outputs to stabilize the robot's trunk along a given reference trajectory, we simply can reuse these control outputs within the model of our Kalman filter without re-solving another optimization problem. A key component of our algorithm is that due to our learning module described in Sec. \ref{learning}, we do not explicitly require a contact detection algorithm (i.e., $n_{k}^{c}$ can be received from the reference instead of explicitly measuring it), as errors associated with our Kalman filter are corrected by the second step described by:
\begin{equation}
\label{eq:GRU}
\begin{aligned}
    \mathbf{O}_{k} = \delta(&\hat{\mathbf{x}}_{k-N:k}, \mathbf{L}_{k-N:k}^{\rm depth}, 
    \mathbf{p}_{k-N:k},
    \dot{\mathbf{p}}_{k-N:k}, \\
    &\mathbf{a}^{\rm imu}_{k-N:k},
    \boldsymbol{\alpha}^{\rm imu}_{k-N:k}, 
    \mathbf{f}_{k-N:k})
\end{aligned}
\end{equation}
where $\delta$ is a function describing the GRU (see Sec. \ref{learning}), with the following inputs: $\hat{\mathbf{x}}_{k-N:k}$, which is the estimated trunk state from \eqref{eq:kalman}, $\mathbf{L}_{k-N:k}^{\rm depth}$ is the latent space of the depth image from our ViT described in Sec. \ref{learning}, $\mathbf{p}_{k-N:k}$ and $\dot{\mathbf{p}}_{k-N:k}$ are the estimated footstep positions and velocities (obtained from the joint encoder positions and velocities, as explained in Sec. \ref{measurement}), $\mathbf{a}^{\rm imu}$ and $\boldsymbol{\alpha}^{\rm imu}$ are the IMU linear and angular accelerations respectively (we use these IMU accelerations because they may indicate instances of slipping or sudden falling),
$\mathbf{f}_{k-N:k}$ are the same ground reaction forces described in \eqref{eq:kalman}, the GRU output is given by $\mathbf{O}_k=[\bar{\mathbf{x}}_{k}, \boldsymbol{\mu}_{k}]$, where $\bar{\mathbf{x}}_{k}$ is the updated trunk state and $\boldsymbol{\mu}_{k}$ is the absolute error between the GRU prediction and ground truth ($\mathbf{x}^{\rm mocap}$) written as $|\bar{\mathbf{x}}_{k}-\mathbf{x}^{\rm mocap}_{k}|$, and we use $N$ time steps to indicate the recurrent nature of the GRU. We include ground reaction forces as inputs as they contain information on the reference trajectory and contact sequence from the MPC optimization, and depth images to convey robot height and environmental semantics in our training procedure.

The main motivation for our use of the Kalman filter within our learning algorithm is the following: (1) The filter offers a reasonable initial estimation of the state, serving as a warm-start for our network and leverages the received domain knowledge, useful for generalization. We substantiate this claim through an ablation study in which we exclude the Kalman filter state as input to the GRU, resulting in a noticeable decline in overall performance (refer to Section \ref{Results}); (2) As we also acquire knowledge of the uncertainty associated with the GRU prediction, we can identify estimates that may carry significant errors. In such instances, users can rely on the Kalman filter estimate, as it remains unaffected by any learning components.

\begin{table}[]
    \centering
    \captionsetup{format=plain, font=small}
        \caption{Notation of Critical variables. Frames are defined in either world frame (W), body frame (B), or neither (N/A)}
\begin{tabular}{|c|c|c|c|}
\hline Name & Description & Size & Frame \\
\hline 
${\mathbf{x}}$ & Trunk State from System Model& $\mathbb{R}^{12\times 1}$  & W\\
$\hat{\mathbf{x}}$ & Trunk State from KF& $\mathbb{R}^{12\times 1}$  & W\\
$\bar{\mathbf{x}}$ & Trunk State from GRU& $\mathbb{R}^{12\times 1}$  & W\\
${\boldsymbol{\mu}}$ & Trunk State GRU Prediction Error& $\mathbb{R}^{12\times 1}$  & W\\
$\boldsymbol{\theta}$ & Trunk Orientation & $\mathbb{R}^{3\times 1}$  & W\\
$\mathbf{r}$ & Trunk Position & $\mathbb{R}^{3\times 1}$  & W\\
$\boldsymbol{\omega}$ & Trunk Angular Velocity & $\mathbb{R}^{3\times 1}$  & W\\
$\mathbf{v}$ & Trunk Linear Velocity & $\mathbb{R}^{3\times 1}$  & W\\
$\mathbf{p}$, $\dot{\mathbf{p}}$ & Footstep Position, Velocity & $\mathbb{R}^{12\times 1}$  & B\\
$i$ & Foot Number & $\mathbb{R}^{1}$  & N/A\\
$k$ & Time Step & $\mathbb{R}^{1}$  & N/A\\
$n$, $n_{c}$ & $\#$ of Feet, $\#$ of Feet in contact & $\mathbb{R}^{1}$  & N/A\\
$C^{i}$ & Reference Contact of Foot i & $\mathbb{R}^{1}$  & N/A\\
$\mathbf{f}^{i}$ & Ground Reaction Force & $\mathbb{R}^{3\times 1}$  & W\\
$\boldsymbol{\Theta}^{i}$, $\dot{\boldsymbol{\Theta}}^{i}$ & Joint Encoder Position, Velocity & $\mathbb{R}^{3\times 1}$  & N/A\\
$\mathbf{L}^{\rm depth}$ & Latent Space of Depth Image & $\mathbb{R}^{128\times 1}$  & N/A\\
$\mathbf{I}^{\rm depth}$ & Raw Depth Image & $\mathbb{R}^{224\times 224}$  & N/A\\
$\hat{\mathbf{I}}$ & Inertia Tensor of Trunk & $\mathbb{R}^{12\times 12}$  & W\\
${\mathbf{I}_{n}}$ & Identity Matrix & $\mathbb{R}^{n\times n}$  & N/A\\
\hline
\end{tabular}
    \label{tab:my_label}
\end{table}

\subsection{Measurements}
\label{measurement}
For our Kalman filter described in Sec. \ref{kalman_filter}, we use IMU measurements that provide the Euler angles of the trunk (converted from raw quaternion) $\boldsymbol{\theta}^{\rm imu} = [\theta_x, \theta_y,\theta_z]^{\top}$ (roll, pitch, and yaw), and corresponding angular velocity 
$\boldsymbol{\omega}^{\rm imu}=[\omega_x, \omega_y, \omega_z]^{\top}$. To measure the linear velocity of the trunk we first convert joint encoder position ($\boldsymbol{\Theta}$) and velocity ($\dot{\boldsymbol{\Theta}}$) using the Jacobian ($\mathbf{J}$) into footstep velocities or $\dot{\mathbf{p}} = \mathbf{J}(\boldsymbol{\Theta})\dot{\boldsymbol{\Theta}}$. Leg odometry can then measure linear velocity of the trunk, $\mathbf{v}^{\rm odom}=[v_x, v_y, v_z]^{\top}$, and robot height, $r_{z}^{\rm odom}$, using the following relationship:
\begin{align}
    \label{eq:Kinematic}
    \mathbf{v}^{\rm odom}=-\frac{1}{n_c} \sum_{i=1}^{n}\mathbf{R}_{b}^{w} (\dot{\mathbf{p}}^{i}+\boldsymbol{\omega}^{\rm imu} \times \mathbf{p}^{i}) \text{ (foot $i$ in contact)}
\end{align}

\begin{align}
    \label{eq:Kinematic2}
    r_{z}^{\rm odom}=-\frac{1}{n_{c}}\sum_{i=1}^{n}p_{z}^{i} \text{ (foot $i$ in contact)}
\end{align}
where $n$ is the number of feet, and $n_c$ is the number of feet in contact with the ground, and $p_{z}$ is the height component of the footstep (received from forward kinematics of joint encoder position), and $\mathbf{R}_{b}^{w}$ is the rotation matrix from body to world frame. Note, this measurement assumes no foot slippage. Lastly, for the training module described in Sec. \ref{learning}, we will measure motion capture data symbolized by $\mathbf{x}^{\rm mocap}$, and the raw depth image as $\mathbf{I}^{\rm depth}$.   




\subsection{Model}
\label{model}
The model used for our Kalman filter is based on the single-rigid body that is subject to forces at a contact patch. While ignoring leg dynamics and assuming small angle approximations in roll and pitch \cite{forceMPC} does simplify the actual robot dynamics, we argue that the decreased computational cost outweighs this issue. The added error due to the model's simplification can also be accounted for by the introduction of our learning module discussed in Sec. \ref{learning}. Simplifying this model's nonlinear dynamics through approximations \cite{forceMPC} and expressing it in discrete-time form yields:
\begin{subequations}
\begin{align}
\label{eq:model}
    \mathbf{x}_{k+1}=\mathbf{Dyn}(\mathbf{x}_{k},\mathbf{f}_{k})
\end{align}
\label{eq:stance_model}
\begin{align} 
    \mathbf{Dyn}&(\mathbf{x}_{k},\mathbf{f}_{k})=
    \begin{bmatrix} 
    \boldsymbol{\theta}_{k} + \Delta T \mathbf{R}_{b}^{w\top}\boldsymbol{\omega}_{k} 
    \\
    \mathbf{r}_{{\rm}k} + \Delta T \mathbf{v}_{{\rm}k}
    \\
    \boldsymbol{\omega}_{k}+\Delta T(\sum_{i=1}^{n}\hat{\mathbf{I}}^{-1}\left[\mathbf{p}^{b}_{i,k}\right]_{\times}\mathbf{f}^{i}_{k})
    \\
    \mathbf{v}_{k}+\Delta T(\sum_{i=1}^{n}\frac{\mathbf{f}^{i}_{k}}{m}+\mathbf{g})
    \end{bmatrix}
\end{align}
\end{subequations}
where $\mathbf{x}_{k}$ is the trunk state of the robot, the ground reaction forces are represented by $\mathbf{f}_{k}^{i}$ for each foot $i$, $m$ is the total mass of the robot, $\mathbf{g}$ is the gravity vector, $\mathbf{\hat{I}}$ is the inertia tensor matrix in world coordinates, $\mathbf{p}_{i,k}^{b}$ is the footstep position of foot $i$ in the body frame, $\Delta T$ is the discretized time step, and lastly, $[\bullet]_\times$ is defined as the skew-symmetric matrix (same definition as used in \cite{forceMPC}). We can rewrite \eqref{eq:stance_model}:
\begin{align}
\label{eq:model}
    \mathbf{x}_{k+1}=(\mathbf{I}_{12} + \mathbf{A}_{k}\Delta T)\mathbf{x}_{k} + (\mathbf{B}_{k}\Delta T)\mathbf{f}_{k} + \mathbf{g}\Delta T
\end{align}
where $\mathbf{I}_{12}\in\mathbb{R}^{12 \times 12}$ is an identity matrix (not to be confused notationally with the inertia tensor matrix $\hat{\mathbf{I}}$), $\mathbf{f}_{k}\in\mathbb{R}^{3n \times 1}$ is the ground reaction force vector (consisting of the $x$, $y$, and $z$ components of force), stacked vertically for each of the $n$ feet, and $\mathbf{g}\in\mathbb{R}^{12 \times 1}$ is the gravity vector, or $\mathbf{g}=[0,0,0,0,0,0,0,0,0,0,0,-9.81]^\top$, and $\mathbf{A}_{k}\in\mathbb{R}^{12 \times 12}$ and $\mathbf{B}_{k}\in\mathbb{R}^{12 \times 3n}$ represent the discrete-time system dynamics:
\begin{equation}
\label{A_matrix}
\mathbf{A}_{k}=\left[\begin{array}{cccc}
\mathbf{0}_3 & \mathbf{0}_3 & \mathbf{R}_{b,k}^{w\top} & \mathbf{0}_3 \\
\mathbf{0}_3 & \mathbf{0}_3 & \mathbf{0}_3 & \mathbf{I}_3 \\
\mathbf{0}_3 & \mathbf{0}_3 & \mathbf{0}_3 & \mathbf{0}_3 \\
\mathbf{0}_3 & \mathbf{0}_3 & \mathbf{0}_3 & \mathbf{0}_3
\end{array}\right]
\end{equation}
\begin{equation}
\label{B_matrix}
\mathbf{B}_{k}=\left[\begin{array}{ccc}
\mathbf{0}_3 & . . . & \mathbf{0}_3 \\
\mathbf{0}_3 & . . . & \mathbf{0}_3 \\
\hat{\mathbf{I}}^{-1}\left[\mathbf{p}^{b}_{1,k}\right]_{\times} & . . . & \hat{\mathbf{I}}^{-1}\left[\mathbf{p}^{b}_{n,k}\right]_{\times} \\
\mathbf{I}_3/m & . . . & \mathbf{I}_3/m
\end{array}\right]
\end{equation}
During operation of the robot, the MPC QP outputs ground reaction forces of the feet in contact with the ground, which are assigned into joint motor torques. These outputs enable the robot trunk to follow a reference state over a time horizon \cite{forceMPC}. Thus, we can simply use these forces with the trunk state and footstep positions at time step $k$ in the model described by \eqref{eq:stance_model}, to predict the trunk state at time step $k+1$.


\subsection{Kalman Filter}
\label{kalman_filter}
From Sections \ref{measurement} and \ref{model}, we have the necessary information to use the Kalman filter equations, which are:
\subsubsection{Prediction}
\begin{align}
    \hat{\mathbf{x}}_{k|k-1}
    & = 
    \mathbf{Dyn}(\mathbf{x}_{k-1|k-1},\mathbf{f}_{k-1|k-1}) \\
    \mathbf{P}_{k|k-1} 
    & = 
    \mathbf{F}_{k}\mathbf{P}_{k-1|k-1}\mathbf{F}_{k}^\top + \mathbf{Q} 
\end{align}
\subsubsection{Update}
\begin{align}
    \hat{\mathbf{y}}_{k} 
    &= 
    \mathbf{z}_{k} - \mathbf{H}\hat{\mathbf{x}}_{k|k-1} \\
    \mathbf{S}_{k}
    & = 
    \mathbf{H}\mathbf{P}_{k|k-1}\mathbf{H}^\top + \mathbf{R} \\
    \mathbf{K}_{k}
    &=
    \mathbf{P}_{k|k-1}\mathbf{H}\mathbf{S}_{k}^{-1} \\
    \hat{\mathbf{{x}}}_{k|k}
    & = 
    \hat{\mathbf{x}}_{k|k-1} + \mathbf{K}_{k}\hat{\mathbf{y}} \\
    \mathbf{P}_{k|k}
    &=
    (\mathbf{I}_{12}-\mathbf{K}_{k}\mathbf{H})\mathbf{P}_{k|k-1}
\end{align}
where $\mathbf{F}_{k}=e^{(\mathbf{A}_{k}\Delta T)}$, $\mathbf{z}_{k}\in\mathbb{R}^{10 \times 1}$ are the measurements received directly from the heading angles of the IMU and odometry (see Sec. \ref{measurement}) written as $\mathbf{z}_{k}=[\boldsymbol{\theta}^{\rm imu}_{k},r_{k,z}^{\rm odom},\boldsymbol{\omega}^{\rm imu}_{k},\mathbf{v}^{\rm odom}_{k}]^\top$. Thus, $\mathbf{H}\in\mathbb{R}^{10 \times 12}$, selecting the components of $\hat{\mathbf{x}}_{k}$ that match the components of $\mathbf{z}_{k}$, or all components except for $r_x$ and $r_y$ horizontal trunk position.


\subsection{Improving Estimation through Learning}
\label{learning}
\begin{figure}[!t]
    \centering
    \includegraphics[width=0.95\columnwidth]{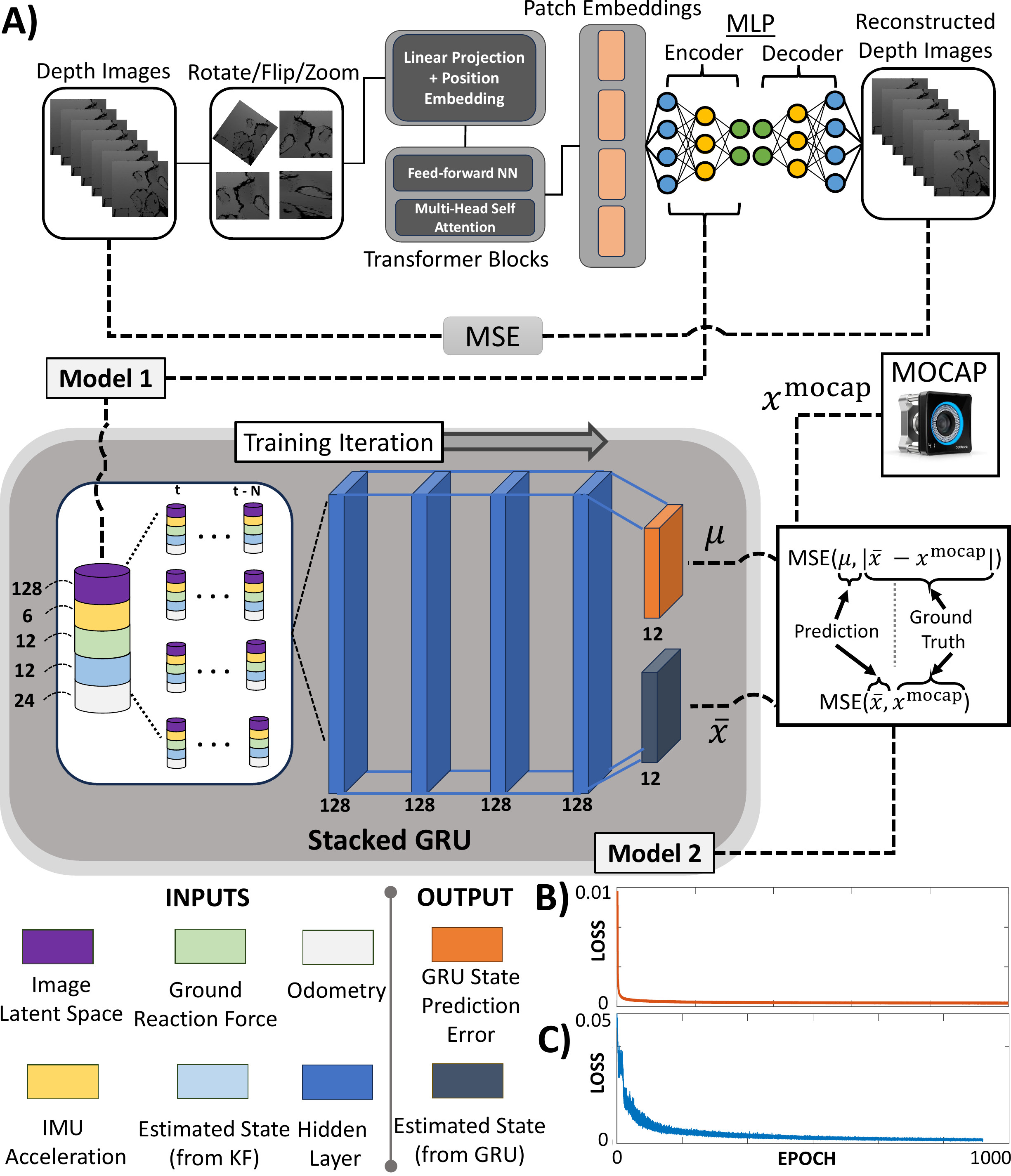}
    \captionsetup{font=small} 
    \caption{Transformer and Gated Recurrent Unit (GRU) network architecture. From  $\mathbf{(A)}$, model 1 is the transformer model, model 2 is the GRU ($\delta$) that predicts the robot's trunk state and uncertainty of its own prediction. Input/output state and hidden layer sizes indicated by the numbers. Training loss of model 1 shown in ($\mathbf{B}$) and for model 2 in ($\mathbf{C}$). MSE is the loss function (see Sec. \ref{learning}). }
    \label{networks}
\end{figure}

A ML approach is employed to consider nonlinearities associated with model and measurement error to better estimate the state of the robot's trunk (see Fig. \ref{networks}). The approach includes training a ViT and then using the ViT model to train a GRU offline, before employing them simultaneously in the testing phase online. We first collect the following data used for our learning modules (the robot moved randomly in all terrains except for slippery and inclined surfaces, where it was commanded to move straight): the ground truth state of the robot's trunk using a motion capture system or $\mathbf{x}^{\rm mocap}$, joint positions and velocities, IMU data or $\mathbf{a}^{\rm imu}$ and $\boldsymbol{\alpha}^{\rm imu}$, depth images or $\mathbf{I}^{\rm depth}$, control outputs of our MPC controller, or $\mathbf{f}$ (see Sec. \ref{model}), and foot contact reference matrix $\mathbf{C}$. We can then make use of \eqref{eq:kalman} and the data collected to calculate the output of the Kalman filter, or $\hat{\mathbf{x}}$, at each time step during collection. We use an autoencoder to convert the raw depth image $\mathbf{I}^{\rm depth}$ to the latent space $\mathbf{L}^{\rm depth}$ through a ViT architecture similar to \cite{MASKED_ViT}. By applying various operations such as rotation, flipping, and zooming on the collected depth images (to facilitate generality of the images for training), we convert these depth images into grayscale and then input them to the encoder. The encoder consists of a patch size equal to $16$, embedding dimension of $128$, depth of $4$, multi-layer perception ratio of $4$, and number of heads equal to $4$. The decoder consists of these same hyperparameters. The AdamW \cite{loshchilov2019decoupled} optimizer is used with weight decay of $0.1$, learning rate of $4e^{-4}$, and $\beta_{1}=0.90$, $\beta_{2}=0.95$. The Mean Squared Error (MSE) loss function is used to minimize the error between the original depth image, and the reconstructed depth image from the output of our decoder. After training, we use the encoder output to represent the latent space of our depth image ($\mathbf{L}^{\rm depth}$). 

We then use \eqref{eq:GRU} to predict the state of the robot's trunk using the GRU ($\delta$), which requires the Kalman filter estimate of the trunk, latent space of the depth image, odometry (i.e., $\mathbf{p}$ and $\dot{\mathbf{p}}$), IMU, and ground reaction forces as input, over a receding horizon of $N$ time steps (we chose $N=10$). $\delta$ also outputs the error between the predicted state output and the ground truth, or $\boldsymbol{\mu}=|\bar{\mathbf{x}}-\mathbf{x}^{\rm mocap}|$, which is learned during training. In other words, $\boldsymbol{\mu}$ helps provide an indicator of uncertainty of our GRU model. This uncertainty may be useful in a stochastic control/planning framework, or used as an indicator to employ the estimate from the Kalman filter instead of the GRU model in certain cases (i.e., $\hat{\mathbf{x}}$ instead of $\bar{\mathbf{x}}$). The GRU uses a learning rate of $1e^{-5}$, number of hidden layers of $4$ with each layer of size $128$, batch size of $64$, Adam optimizer with weight decay of $1e^{-5}$, and with MSE as our loss function. We also normalize the input and output data from $0$ to $1$ (using min/max of dataset) to help stabilize the training. See $\mathbf{(B)}$ and $\mathbf{(C)}$ in Fig. \ref{networks} for the training loss of the autoencoder and GRU respectively. 

\section{Experimental Validation}
\begin{figure*}[!t]
    \centering
    \includegraphics[width=6.8in]{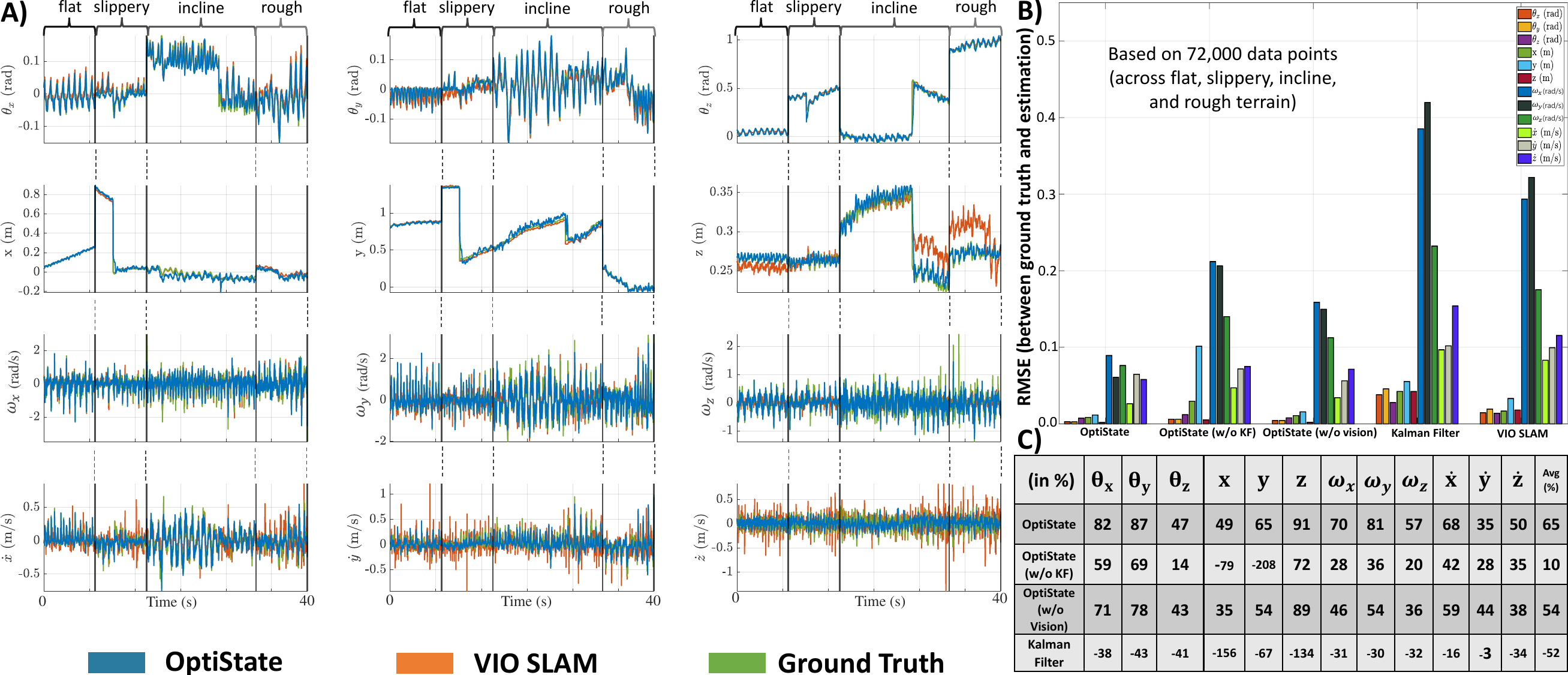}
    \captionsetup{font=small} 
    \caption{Results during the online testing phase, as described in Sec. \ref{Results}. In $\mathbf{(A)}$ we show the state estimation for all state components from OptiState, VIO SLAM, and the ground truth. We show 4 distinct trajectories connected by solid lines to symbolize the various terrain under evaluation, such as flat, slippery, incline, and rough terrain. The RMSE results over all 4 trajectories and per state component are shown in $\mathbf{(B)}$, and includes OptiState without the Kalman filter input, or vision input, and the Kalman filter alone. Lastly, we show the percentage improvement of RMSE over the VIO SLAM baseline for each state in $\mathbf{(C)}$ per estimation algorithm shown in the first column.}  
    \label{results}
\end{figure*}

\begin{figure}[!t]
    \centering    \includegraphics[width=0.99\columnwidth]{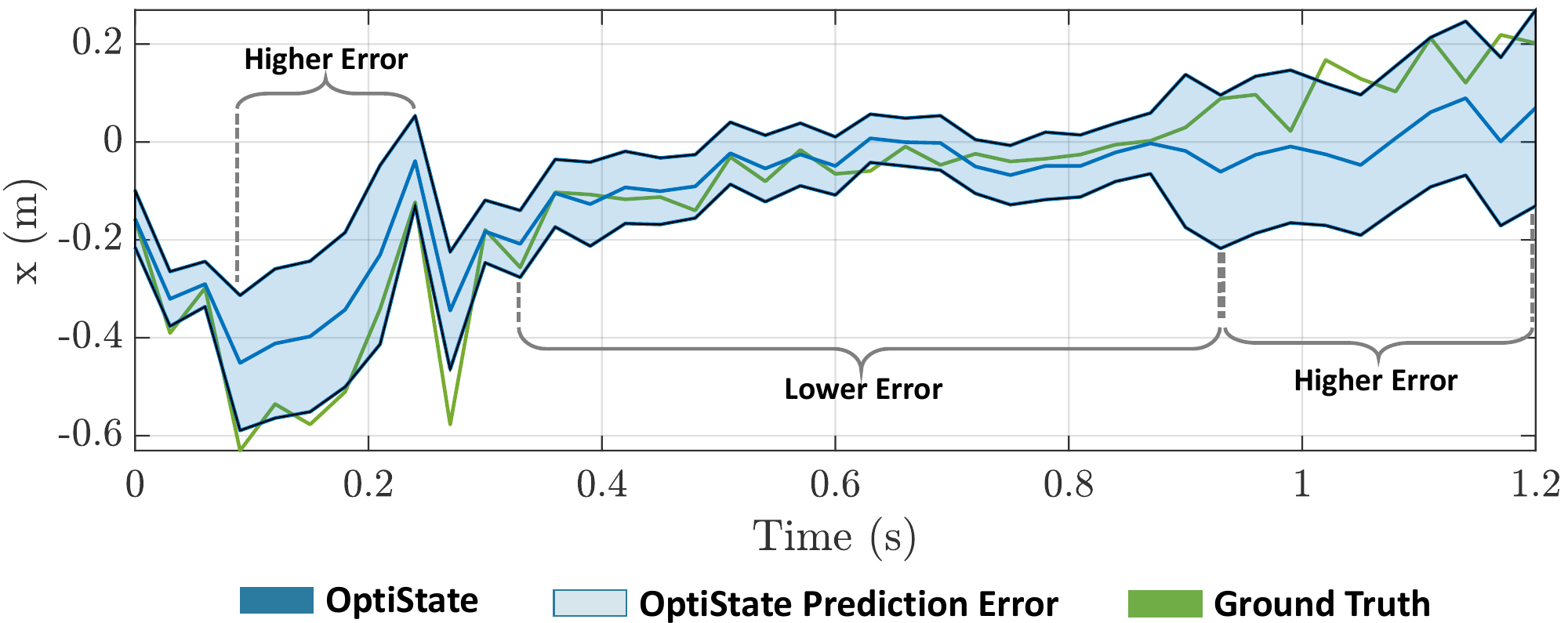}
    \captionsetup{font=small} 
    \caption{Example of predicting the uncertainty of the GRU's (OptiState) $x$ position, or $\mu_{x}$. The shaded blue represents $\bar{{x}}\pm\mu_{x}$.} 
    \label{predict_error}
\end{figure}
For data collection, we used six VICON Vero v2.2 cameras running at 330 Hz to estimate the ground truth trunk state of our SCALER robot \cite{tanaka2022}. We compare OptiState with VIO SLAM from the the Intel RealSense T265 camera \cite{hausamann2021evaluation} running at 200 Hz. For collecting depth images, we use the Intel RealSense D435 camera pointed at a 45$\degree$ angle to the ground, running at 60 Hz. Our encoder runs up to 500 Hz, which is well above the frequency of our camera. Note, our GRU estimator (i.e., up to 1600 Hz) is limited by the frequency of our motion capture system (i.e., 330 Hz), where lower frequency components are simply repeated during the training procedure. Evaluation was done on an Intel i7 8850H CPU  with a Quadro 3200 GPU. As the T265 camera is our main baseline, and to compare this baseline with our architecture, we down-sample our GRU estimation to 200 Hz, with $\Delta T$=0.005 s when plotting results. 

\subsection{Results}
\label{Results}
Overall, we collected 16 different trajectories for training the GRU using a trot gait, where each trajectory lasted between 1 to 2 minutes (total of $\approx$ 288,000 data points) and included flat, slippery, incline, and rough terrain. See Fig. \ref{intro} and accompanied video for the robot motion during collection. After training, we evaluate on four trajectories previously unseen during training (one trajectory per surface). The results of the testing phase are shown in \textbf{(A)}, \textbf{(B)} and \textbf{(C)} of Fig. \ref{results}. Note that in \textbf{(A)}, we display only a portion of the four trajectories and baseline comparison for easier visibility. However, in \textbf{(B)}, we computed the Root Mean Squared Error (RMSE) between the ground truth and the estimation algorithm over the entire evaluation testing set, which totals 72,000 data points at $\Delta T$=0.005 s. In \textbf{(A)}, we see that both OptiState and VIO SLAM performed well, though VIO SLAM struggled more with predicting linear velocity and, notably, the robot's $z$ height which becomes apparent during the incline and especially on rough terrain. This may be 
due to the fact that our rough terrain contains compliant surfaces, which can easily disrupt estimation systems and produce drift particularly in $z$ height.

In \textbf{(B)}, we compared our estimation algorithm, OptiState, under three different conditions: one without the Kalman filter state $\hat{\mathbf{x}}$ as input to the GRU, another without the vision component or $\mathbf{L}^{\rm depth}$ as input to the GRU, and using the Kalman filter alone as described in Sec. \ref{kalman_filter}. We also included a baseline comparison with the VIO SLAM method (T265). In \textbf{(C)}, we compared RMSE percent improvements over our VIO SLAM baseline for each state component, with the average percentage across all state components shown in the right-most column. OptiState's percent improvement was generally higher by not excluding the Kalman filter state or depth image input, except for $\dot{\mathbf{y}}$, confirming the importance of including these inputs into our GRU. OptiState outperformed both the Kalman filter and VIO SLAM for all components. The Kalman filter had the poorest performance at average RMSE improvement of -52$\%$, while OptiState performed the best at 65$\%$. The Kalman filter's errors may be due to slippery and rough terrain conditions, which can violate the model's assumptions (see Sec. \ref{model}). However, the Kalman filter showed less error in velocities. Note that during evaluation, the trace of $\mathbf{P}$ (covariance matrix) of the Kalman filter never diverged. The deviation between trace values throughout the entire testing trajectories remained within a small range, always less than $1e^{-5}$. In Fig. \ref{predict_error}, we show one example on how $\boldsymbol{\mu}$, which predicts the absolute error between the GRU estimate and the ground truth, can highlight potential prediction errors. It correctly predicts higher uncertainty when the GRU estimate is less accurate. 
\section{Limitations and Conclusion}
In this work, we demonstrated state estimation of legged robots using joint encoder information, IMU measurements, a ViT autoencoder, and outputs of a Kalman filter that reuses the control forces from MPC in its system model, wrapped within a GRU framework. In addition to predicting the trunk state of the robot, we also showed that we can predict the uncertainty or error of the GRU's prediction. Although OptiState showed improved performance for all states compared to VIO SLAM, there are several limitations to using our approach. First, although the GRU can reach up to 1600 Hz, we must note that the quality and frequency of our estimation is predicated on the ground truth setup (i.e., in our setup, up to 330 Hz). Because we employed a motion capture system for ground truth with a limited training space, accurate prediction of $x$ and $y$ world coordinate positions is challenging when the robot goes beyond this space---although velocity components can be predicted normally. In these cases, the user may integrate the velocity components of the GRU predictions to get a more accurate position, although accumulation of drift would occur. Overall, we have created a robust state estimation system for legged robots using a hybrid ML-Kalman filter approach, and emphasize ground truth motion capture collection as a key area of future improvement. 

\bibliographystyle{IEEEtran}
\bibliography{bibliography}

\end{document}